\documentclass[10pt,twocolumn,letterpaper]{article}

\usepackage{cvpr}
\usepackage{times}
\usepackage{epsfig}
\usepackage{graphicx}
\usepackage{amsmath}
\usepackage{amssymb}
\usepackage{multirow}
\usepackage{bbding}
\usepackage{pifont}
\usepackage[pagebackref=true,breaklinks=true,letterpaper=true,colorlinks,bookmarks=false]{hyperref}

\cvprfinalcopy % *** Uncomment this line for the final submission

\def\cvprPaperID{4523} % *** Enter the CVPR Paper ID here

\newcommand{\cmark}{\ding{51}}%
\ifcvprfinal\fi
\begin{document}
	
	%%%%%%%%% TITLE
	\title{Context-Aware Dynamic Feature Extraction for 3D Object Detection in Point Clouds}
	
	\author{Yonglin Tian, Lichao Huang, Xuesong Li, Kunfeng Wang, Zilei Wang, Fei-Yue Wang}
		%\and

	\maketitle
	%\thispagestyle{empty}
	
	%%%%%%%%% ABSTRACT
	\begin{abstract}
		Varying density of point clouds increases the difficulty of 3D detection. In this paper, we present a context-aware dynamic network (CADNet) to capture the variance of density by considering both point context and semantic context. Point-level contexts are generated from original point clouds to enlarge the effective receptive filed. They are extracted around the voxelized pillars based on our extended voxelization method and processed with the context encoder in parallel with the pillar features. With a large perception range, we are able to capture the variance of features for potential objects and generate attentive spatial guidance to help adjust the strengths for different regions. In the region proposal network, considering the limited representation ability of traditional convolution where same kernels are shared among different samples and positions, we propose a decomposable dynamic convolutional layer to adapt to the variance of input features by learning from local semantic context. It adaptively generates the position-dependent coefficients for multiple fixed kernels and combines them to convolve with local feature windows. Based on our dynamic convolution, we design a dual-path convolution block to further improve the representation ability. We conduct experiments on KITTI dataset and our proposed CADNet achieves good performance on 3D detection task in terms of both precision and speed. Our one-stage detector outperforms SECOND and PointPillars by a large margin and runs at the speed of 30 FPS.
	\end{abstract}
	

	% \fi

	\section{Introduction}
\begin{figure}[tbp]
		\begin{center}
			\includegraphics[scale=0.42]{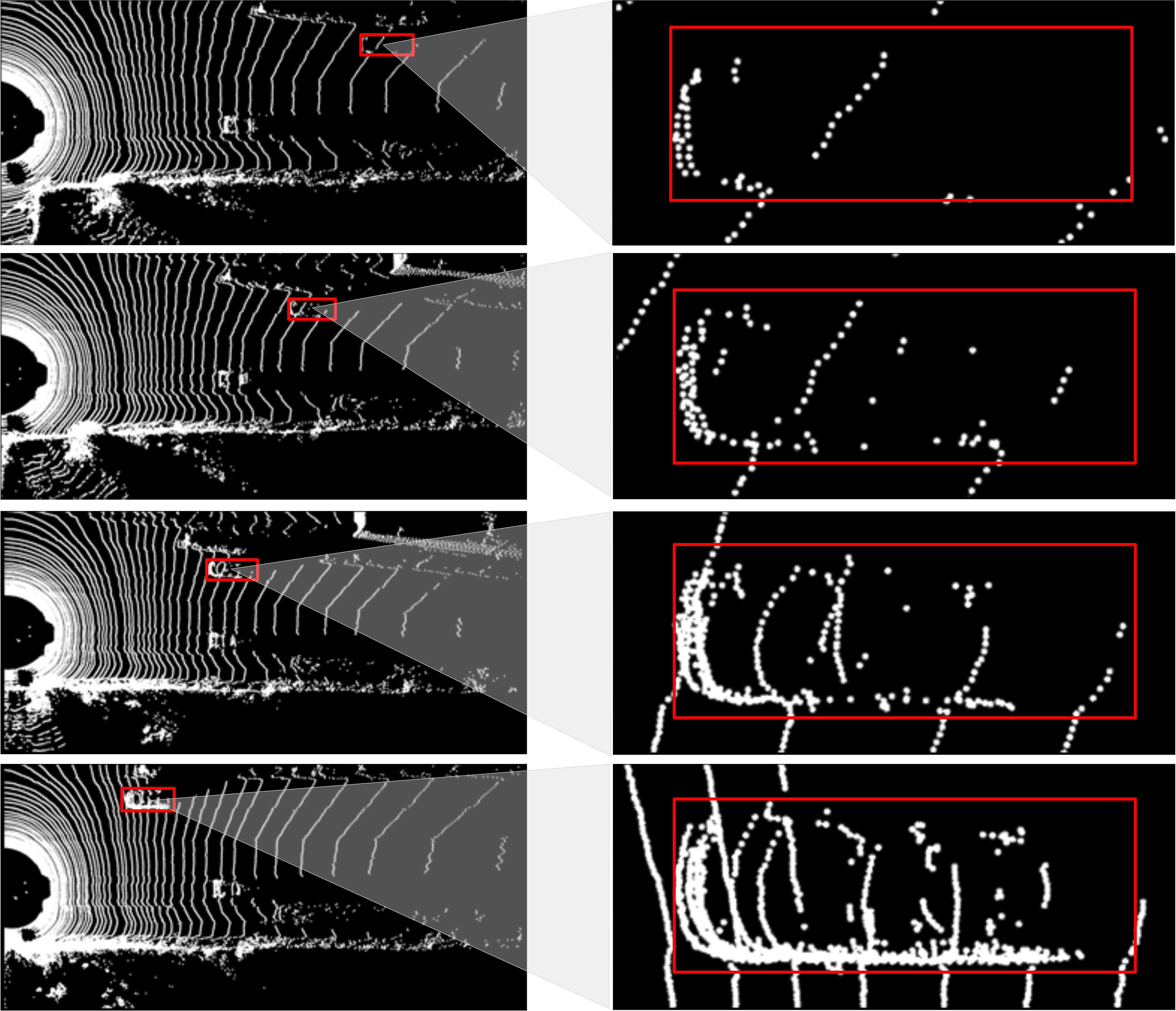}
		\end{center}
		\caption{Illustration of the density variance of the same object in LiDAR scan.}
		\label{fig:varying_density}
	\end{figure}

	Three-dimensional object detection plays a great role in autonomous driving and intelligent transportation systems. LiDAR sensor is one of the key factors to achieve precise localization for 3D detection. However, the varying density of LiDAR points heavily restricts the extraction of features from point clouds. Caused by the position and pose change, even the same object can show different states in LiDAR scans as shown in Fig. \ref{fig:varying_density}. One obvious difference lies in the change of point density. As the object moves away from the sensors, the received LiDAR points become sparser gradually. Such change in density significantly increases the burden of detection from point clouds. Current 3D detectors \cite{chen2017multi,ku2018joint,zhou2018voxelnet,yan2018second,lang2019pointpillars,shi2019pointrcnn} mostly rely on the convolutional neural networks to extract features from point clouds. In traditional convolutional layers, all the positions on the feature maps share the same filters, thereby making the detectors hard to adapt to the varying density. In this paper, we present a new one-stage framework to capture the variance of density in point cloud by designing position-dependent dynamic filters based on local point context and semantic context.
	
	Point clouds have significantly different data format compared with images. They are sparse and unordered in 3D space, making the commonly-used image feature extractors such as VGG \cite{simonyan2014very} and ResNet \cite{he2016deep} hard to be directly deployed in the preprocessing of LiDAR data. To transform point cloud into regular data and process it with existing powerful CNNs, voxelization-based approaches \cite{zhou2018voxelnet,yan2018second,lang2019pointpillars} divide point cloud into voxels or pillars. Then, PointNet \cite{qi2017pointnet,qi2017pointnet++} is applied to generate local features for each small point set and CNNs are used to further process these transformed image-like features. This kind of voxelization eases the difficulty of detection from unordered point cloud and helps to generate regular features in an efficient way. Therefore, voxelization-based feature extraction has become the main component of many popular 3D detection backbones \cite{zhou2018voxelnet,yan2018second,lang2019pointpillars,wang2019voxel,shi2019part}. In this paper, we extend current voxelization method to generate point-level context surrounding each voxelized pillar. It aggregates point features in a wider range and can provide subsequent layers with a much larger receptive field. Therefore, context features are potential to capture the density of local region and can be used for the generation of guidance map to indicate the importance of different areas.   
	
	With voxelized dense features, many works \cite{lang2019pointpillars,yan2018second,zhou2018voxelnet} have utilized CNNs to make final predictions. Although CNNs have achieved great success in many tasks of computer vision and pattern recognition, making precise prediction for objects with heavily changing appearances is still hard \cite{jaderberg2015spatial}. For point clouds, the number of LiDAR points reflected from the same object can decrease from a few hundred to several as the object moves away, which leads to different point densities on the different areas in the LiDAR scan. Because convolutional layers share the same filters among all the positions of the input, it's tough for CNN to adapt to changing features. Dynamic filter network (DFN) \cite{jia2016dynamic} proposes to handle such changes with adaptive filters learned by an extra network and has been successfully applied in many tasks, such as depth completion \cite{tang2019learning} and image classification \cite{chen2019you}. However, replacing traditional convolutional filters with corresponding dynamic ones will greatly increase the number of parameters. To reduce the computation and capture the variance in feature maps, we propose a new convolutional layer named decomposable dynamic convolution (DD-Conv) that adaptively generates the coefficients of multiple kernels and combines them to extract local features. We deploy our DD-Conv at the last layer of each block to respond to high-level semantic context of each position on the feature maps and adapt to the variance of density. Based on our DD-Conv, we build a dual-branch dynamic region proposal network (RPN). These two branches have the same architecture but different weights. With several layer stacked, these branches can better model the variance on the feature maps. This dual-branch design enhances the representation ability of changeable features.
	
	This paper focuses on the extraction of robust features from point clouds with variable density and our main contributions can be summarized as follows:
	\begin{itemize}
		\item We propose a context-aware one-stage 3D detection network to improve the robustness of feature extractor to the variances of point density.
		\item We extend the voxelization method to capture the point context in a wider range which provides the following network with a larger receptive field.  
		\item We propose a decomposable dynamic layer and a dual-branch RPN to adapt to the variances in feature maps by considering the semantic context for each local area.
	\end{itemize}
	
	\section{Related Work}
	3D object detection attracts much attention in recent years. According to the formats of point cloud, we category related methods into point-based methods, voxelization-based methods and point-voxel methods in the first three paragraphs. In the fourth part, we summarize the related works on dynamic convolution networks.
	\subsection{Point-based 3D Detection}
	PointNet \cite{qi2017pointnet} is the pioneer to directly manipulate raw point clouds with neural networks. PointNet uses per-point processing to extend the dimension of each point and uses the max-pooling to aggregate global features. Based on PointNet, PointNet++ \cite{qi2017pointnet++} proposes sampling layer and grouping layer to extract local features which improve the performance of classification and segmentation. PointNet is efficient and effective to process unordered data and lays the foundation for many 3D detection networks \cite{qi2018frustum,shi2019pointrcnn,yang2019std,qi2019deep}. The main difference among these methods is the generation of proposals. F-PointNet \cite{qi2018frustum} generates frustum-shaped proposals in point cloud with the help of 2D detection and uses PointNet to segment positive points for the regression of 3D bounding box. VoteNet \cite{qi2019deep} gets rid of the dependence on 2D detection and proposes a new proposal generation mechanism with deep Hough voting \cite{hough1959machine}. It claims that voting from seed points helps with the generation of more confident and accurate proposals. ImVoteNet \cite{qi2020imvotenet} further combines image cues with point seeds and improves the 3D detection performance. Point RCNN \cite{shi2019pointrcnn} turns to point segmentation network for the generation of 3D proposals. Different from F-PointNet, Point RCNN directly segments the whole point cloud and produces proposal for each foreground point. For indoor 3D detection, DenseFusion \cite{wang2019densefusion} uses the predicted mask from image features to extract the point set and concatenates the point to each pixel on image feature maps. 3DSSD \cite{yang20203dssd} generates candidate points from point cloud using furthest point sampling method. To improve the point recall, 3DSSD propose a feature-based sampling method and preserve more positive points compared with distance-based sampling.
	\subsection{Voxelization-based 3D Detection}
	Voxelization-based approaches for point cloud processing turn irregular points into an ordered tensor to simplify the extraction of features and make full use of existing CNNs. To avoid the occlusion of foreground objects, many works \cite{chen2017multi,ku2018joint,yang2018pixor,yang2018hdnet} transform point cloud into bird’s eye view (BEV). MV3D \cite{chen2017multi} and AVOD \cite{ku2018joint} propose to slice the point cloud into several layers and divide each layer into small grids. The maximum height in each grid is extracted to generate height maps. Together with density features and image data, they design multi-view networks for 3D detection. Instead of using handcrafted features to represent each small region, VoxelNet \cite{zhou2018voxelnet} proposes a data-driven method using PointNet to learn richer representation for each voxel. To accelerate the convolution of sparse data after voxelization, SECOND \cite{yan2018second} applies a sparse convolution algorithm and achieves real-time inference. To avoid 3D convolution operation and further speed up 3D detection, PointPillars \cite{lang2019pointpillars} replaces voxels with pillars which ignores the partition along z axis and achieves satisfactory accuracy and speed. These methods only consider the single-scale voxelization at the early stage. They need to carefully design the voxel size to achieve the tradeoff between the localization ability and computation cost \cite{ye2020hvnet}. VoxelFPN \cite{wang2019voxel} and HVNet \cite{ye2020hvnet} adopt a multi-scale voxelization strategy and achieve better results. They use different sizes to voxelize the point cloud and get multiple pseudo-image feature maps with different resolution. VoxelFPN \cite{wang2019voxel} and HVNet \cite{ye2020hvnet} focus on the extraction and fusion of multi-scale features to improve the performance of 3D detection. Our voxelization approach differs from the multi-scale voxelization method proposed by VoxelFPN \cite{wang2019voxel} and HVNet \cite{ye2020hvnet} on two aspects. First, our method is technically different from them. They independently voxelize the point cloud with different sizes and the produced pseudo-image features have different resolutions. In our method, the generation of context maps is correlated with pillar maps. We aggregate the point-level context in a wider range for each non-empty position on the pillar maps. The context is centred around the non-empty pillar and both context maps and pillar maps have the same resolution. Secondly, our method differs with them in purpose. VoxelFPN \cite{wang2019voxel} and HVNet \cite{ye2020hvnet} aim to get multiple features while we intend to capture the variance of density in the point cloud and provide context information for our dynamic network. 
	\subsection{Point-voxel Methods on 3D Detection}
	Voxelization-based methods can utilize the traditional convolutional layers to process the point cloud conveniently. However, the voxelization operation involves information loss more or less. Besides, to reduce the burden of detection head, voxelized features usually need to be down-sampled by 2 or 4 times which can affect the localization ability. Compared with voxelization-based methods, detection from raw point cloud can well maintain the 3D information inside the LiDAR data but suffers from complicated processing. Recent works propose to achieve the tradeoff between these two methods by using both voxelized features and raw point cloud. \cite{tian2019adaptive,shi2020points} use voxelized features to generate 3D proposals in the first stage and extract raw point features in the second stage. SA-SSD \cite{he2020structure} maps voxel features to point cloud and uses an auxiliary network to predict the segmentation mask and object center in training phase. This auxiliary network can be removed in the test phase to improve the inference speed. PV-RCNN \cite{shi2020pv} generates key points from point cloud and uses them to extract multi-scale voxel features. It combines voxel and point features for each proposal. The key points greatly reduce the search space of region of interest (RoI) pooling compared with extracting point features from whole point cloud. 
	\subsection{Dynamic Filter Networks}
	DFN \cite{jia2016dynamic} introduces a new framework that can generate dynamic filters depending on the input. It uses an extra network to output the parameters of filters for every position of the input and enables adaptive feature extraction. ECC \cite{simonovsky2017dynamic} extends dynamic filters to graph and demonstrates the effectiveness on point cloud classification. To incorporate more information in the neighbouring regions, LS-DFN \cite{wu2018dynamic} generates weights with larger receptive fields using dynamic sampling convolution. Tang et al. \cite{tang2019learning} propose to take image features as the guidance to generate the weights used for depth completion. To reduce the GPU memory consumption, they factorize the dynamic convolution into channel-wise and cross-channel convolution. Recent works \cite{yang2019condconv,chen2020dynamic} use attention mechanism to dynamically compose the convolutional kernels. They adopt the global pooling over the spatial dimension to generate  channel-wise aggregation and use fully-connected layers to generate the attention over different kernels. We share the similar philosophy with these methods to dynamically generate filters according to the input, but we treat the problem in a different way. Different from the methods \cite{jia2016dynamic,simonovsky2017dynamic,wu2018dynamic,tang2019learning} that directly generate the parameters of the kernel, in our approach, the generation of dynamic kernels is decomposed into the prediction of several coefficients of traditional kernels. We use an auxiliary network to learn these coefficients based on local semantic context and perform position-dependent convolution to adapt to the variance on feature maps. \cite{yang2019condconv,chen2020dynamic} model the input-level dynamics and ignore the variance on different positions of the feature maps, while we focus on the variance in the spatial dimension on the feature maps and generate position-dependent kernels. 
	
	\begin{figure*}[ht]
		\centering
		\includegraphics[scale=0.65]{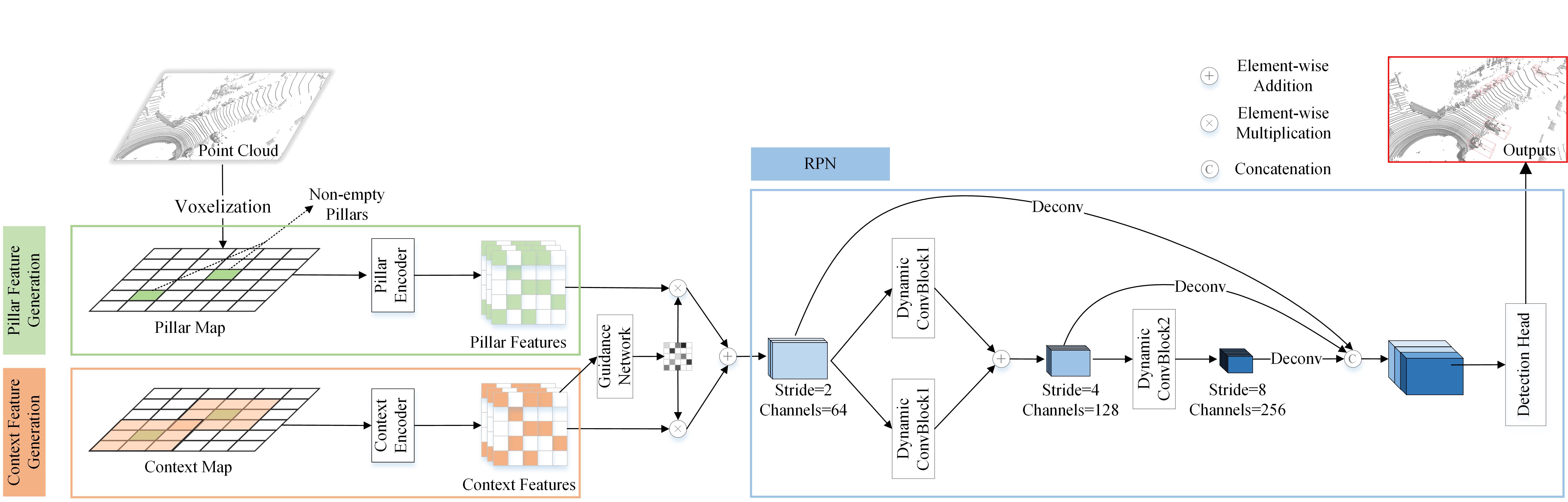}
		\caption{The framework of CADNet. First, the whole point cloud is voxelized into pillars along x and y axes. For non-empty pillars, we extract the point-level contexts within a wider range. Then, pillar encoder and context encoder are used to generate pillar features and the corresponding context features in parallel. The guidance map is produced from context features to focus more attention on valuable regions. We combine the pillar features and context features to feed our dynamic RPN where DD-Conv and a dual-path network are applied to adapt to the variance of input and generate robust features. We use deconvolution and concatenation to fuse multi-level features and 
			use a tiny detection head to output the final detection results.}
		\label{framework}
	\end{figure*}
	\section{Method}
	The framework of our CADNet is illustrated in Fig. \ref{framework}. The main components of our detection network include parallel pillar and context feature extraction network, and the dynamic RPN.
	
	\subsection{Pillar Feature Extraction}
	We first crop the whole point cloud into a cube where the coordinates of each point are ranging in [[$X_{min}$, $X_{max}$],[$Y_{min}$, $Y_{max}$],[$Z_{min}$, $Z_{max}$]]. This helps to exclude some background points and reduce the computation burden. Then, we divide the remaining points into grids along $x$ and $y$ axes with size of $X_{s}$, $Y_{s}$ respectively, and ignore the division along vertical direction like \cite{lang2019pointpillars}. Each cell has the size of $S$,
	\begin{align}
	\begin{split}
	S=[X_{s},\ Y_{s},\ Z_{max}-Z_{min}].
	\end{split}
	\end{align}%
	And the set of all cells can be denoted as:
	\begin{align}
	\begin{split}
	C=\{c_{j} | j=1,...,H*W],
	\end{split}
	\end{align}%
	where $c_{j}$ is the $j$-th cell in the grid. $H$ and $W$ are the height and width of the grid:
	\begin{align}
	\begin{split}
	H=\left \lceil \frac{X_{max}-X_{min}}{X_{s}}\right \rceil,\\
	W=\left \lceil \frac{Y_{max}-Y_{min}}{Y_{s}}\right \rceil,
	\end{split}
	\end{align}%
	where $\lceil \rceil$ denotes the ceil of decimal.
	
\begin{figure}[ht]
		\centering
		\includegraphics[scale=0.7]{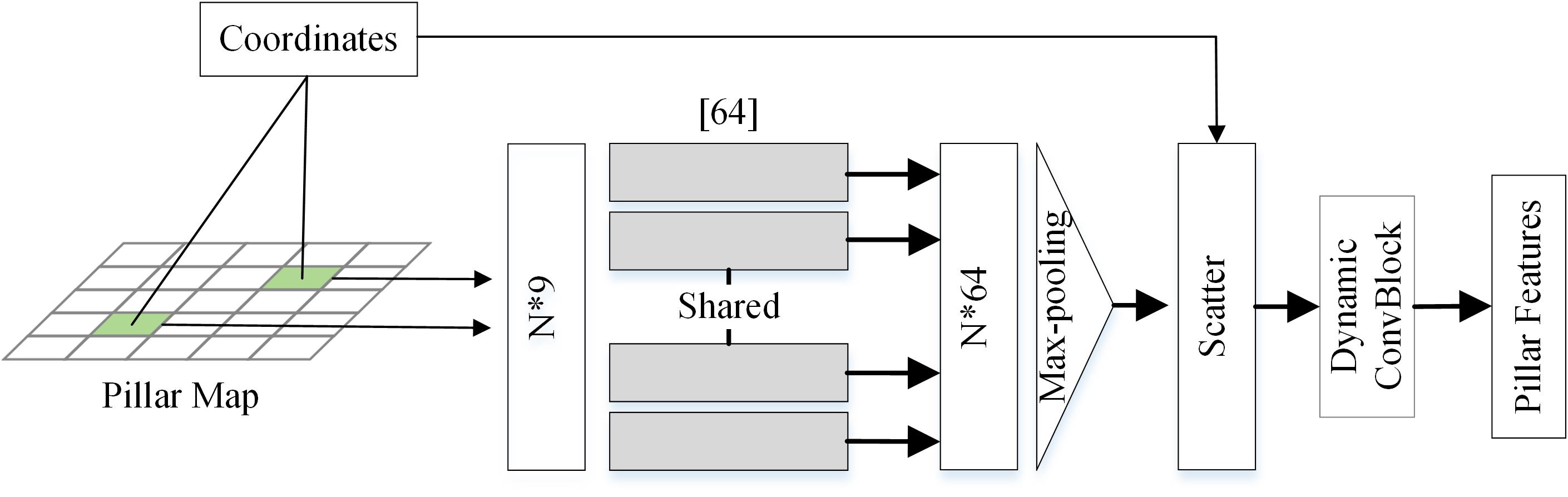}
		\caption{The structure of our pillar encoder.}
		\label{pillar_encoder}
	\end{figure}

\begin{figure}[t]
		\centering
		\includegraphics[scale=0.7]{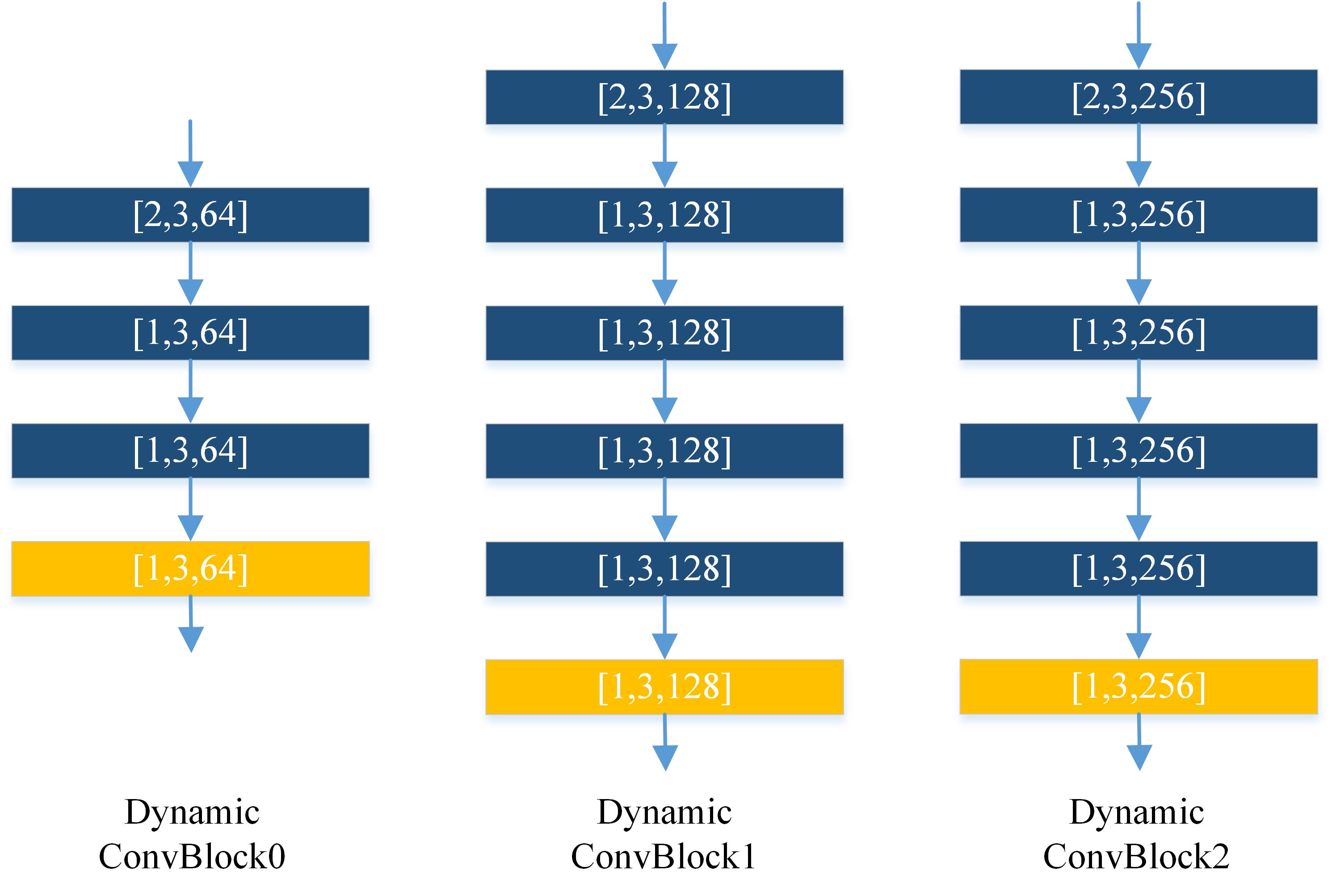}
		\caption{The layouts of our dynamic convolution block. Boxes in blue denote traditional layers and boxes in yellow denote our DD-Conv layers. The numbers in brackets represent stride, kernel size and number of channel in output in turn.}
		\label{ddblock}
	\end{figure}

	To avoid meaningless computation, we only consider non-empty cells if not explicitly specified. For empty cells, we directly pad them to the same size with non-empty ones with zeros in the following processing. We sample $N_{max}$ points for each cell to generate pillars and define the pillar filter to preprocess them. For cell $c_{j}$, we denote its coordinates in BEV as $[coor_{xj},coor_{yj}]$, then the range $R_{pj}$ of corresponding pillar in the cell $c_{j}$ is:
	\begin{align}
	\begin{split}
	R_{pj} = [& [ coor_{xj}-\frac{X_{s}}{2},coor_{xj}+\frac{X_{s}}{2}],\\
	& [coor_{yj}-\frac{Y_{s}}{2},coor_{yj}+\frac{Y_{s}}{2}]].
	\end{split}
	\end{align}
	For points in the range $R_{pj}$, the pillar filter is defined as follows:
	\begin{align}
	\begin{split}
	f_{p}(p_{j}) = \{[x_{ij},y_{ij},z_{ij},x_{ij}^{'},y_{ij}^{'},z_{ij}^{'},x_{ij}^{''},y_{ij}^{''},r_{ij}], i=1,...,N_{max}\},
	\end{split}
	\end{align}
	where $p_{j}$ is the pillar in the position of cell $j$. $N_{max}$ is the number of points in each pillar. $[x_{ij},y_{ij},z_{ij}]$ is the coordinates of point in pillar $p_{j}$ in LiDAR coordinate system. $[x_{ij}^{'},y_{ij}^{'},z_{ij}^{'}]$ is coordinates relative to the mean of all points within the pillar. [$x_{ij}^{''},y_{ij}^{''}$] is coordinates relative to the center of pillar $p_{j}$. The reflectance of $i$-th point in pillar $p_{j}$ is denoted as $r_{ij}$. After the preprocessing, we put the pillar map into our pillar encoder as shown in Fig. \ref{pillar_encoder} to generate pillar features. Firstly, a PointNet \cite{qi2017pointnet} is applied to extract the feature inside each pillar. We use a fully-connected layer to map each point to high dimension and use a max-pooling layer to aggregate the points inside the same pillar. With the coordinates information, we can easily scatter the feature of non-empty pillar back to a dense map. Secondly, our dynamic convolution block is used to extract local features around each pixel on the pillar map. We stack four convolutional layers in this block. All these layer use $3\times3$ kernels to cast input to 64-channel output. The first layer down-samples the input by 2 times with a stride of 2. Only the last layer adopts our DD-Conv to capture the variances on spatial dimension. The layout of Dynamic ConvBlock0 is shown in the left part of Fig. \ref{ddblock}.
	\subsection{Context Feature Extraction}
	To capture the variance of local density, we propose to extract the point-level context from point clouds. The points around each non-empty pillars in a larger range are gathered as shown in Fig. \ref{framework}. We consider the points within the range of 3 times the width and 3 times the length of the pillar. We keep $2 \times N_{max}$ points for each context. 
	The range $R_{tj}$ of the context for cell $c_{j}$ in BEV is:
	\begin{align}
	\begin{split}
	R_{tj}=[& [coor_x-\frac{3X_{s}}{2},coor_{x}+\frac{3X_{s}}{2}],\\
	& [coor_{y}-\frac{3Y_{s}}{2},coor_{y}+\frac{3Y_{s}}{2}]].
	\end{split}
	\end{align}
	The following context filter is used to generate context map. Compared with the pillar filter, we ignore the absolute coordinates of points in LiDAR coordinate system considering they have been included in pillar filter. For every point in the range $R_{tj}$ of Context filter is defined as follows:
	\begin{align}
	\begin{split}
	f_{t}(t_{j}) = \{[x_{i}^{'},y_{i}^{'},z_{i}^{'},x_{i}^{''},y_{i}^{''},r_{i}], i=1,...,2N_{max}\},
	\end{split}
	\end{align}
	where $t_{j}$ is the context in the position of cell $j$ for pillar $p_{j}$. $N_{max}$ is the number of points of each pillar. $[x_{i}^{'},y_{i}^{'},z_{i}^{'}]$ is coordinates relative to the mean of all points within the context. [$x_{i}^{''},y_{i}^{''}$] is coordinates relative to the center of context $p_{j}$.
	
	Our context feature encoder shares similar design with pillar encoder as shown in Fig. \ref{pillar_encoder}, except that the dimension of each point in the context is changed from 9 to 6. Context features aggregate the point features in a wider range and possess a larger receptive field. They are used to generate the guidance maps to indicate the importance of regions in pillar feature maps and context features maps. We simply applied one $1\times1$ covolutional layer to the context feature maps and generate two guidance maps, one for pillar features and the other for context feature maps.
	\subsection{Decomposable Dynamic Convolution}
\begin{figure}
		\begin{center}
			\includegraphics[scale=0.8]{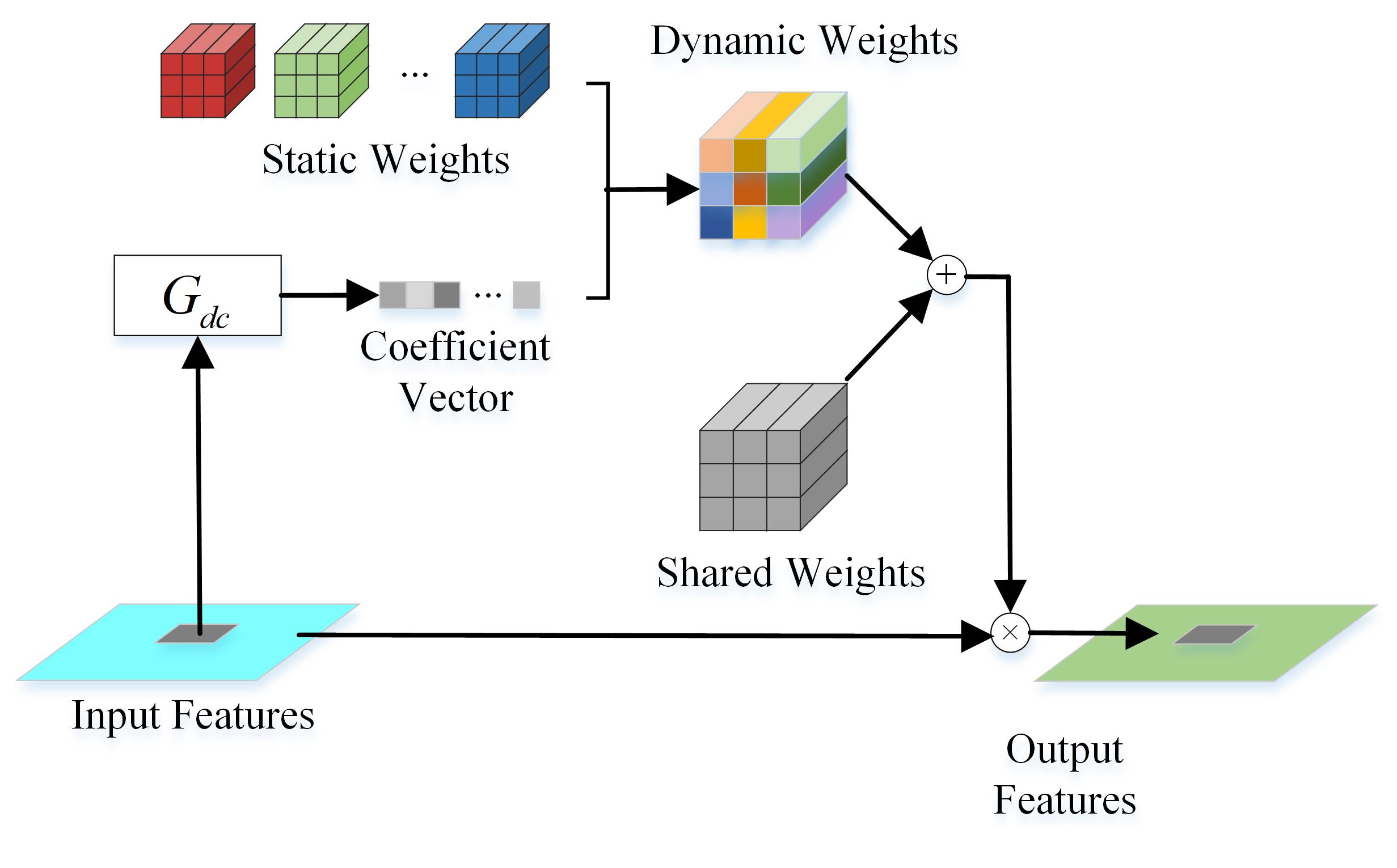}
		\end{center}
		\caption{Illustration of our decomposable dynamic convolution.  }
		\label{fig:dd-conv}
	\end{figure}

	To adapt to the variance in feature maps, we design a decomposable dynamic convolution to adaptively change the weights of convolutional kernels depending on the local input. Traditional convolutional layer takes regular feature maps as inputs which can be denoted as $I \in \mathbb{R}^{h\times w \times c}$ and output new feature maps  $O \in \mathbb{R}^{h^{'}\times w^{'} \times c^{'}}$ using multiple filters with weights $W \in \mathbb{R}^{s\times s \times c \times c^{'}}$. Here, $h$, $w$ and $c$ are the height, width and number of channels of input feature. And $h^{'}$, $w^{'}$ and $c^{'}$ are size of output feature. The kernel size is denoted by $s$. Traditional convolution can be formalized as:
	\begin{align}
	\begin{split}
	O=W \otimes I.
	\end{split}
	\end{align}%
	And our decomposable dynamic convolution is compatible with traditional convolution. It is comprised of two parts, i.e., dynamic weights and shared weights as shown in Fig.\;\ref{fig:dd-conv}. We formulate our DD-Conv as follows:
	\begin{align}
	\begin{split}
	O=(W_{d}(I;\theta_{d}) + W_{s}) \hat{\otimes} I,
	\end{split}
	\end{align}%
	where $\theta_{d}$ is parameters of the network to learn dynamic weights $W_{d} \in \mathbb{R}^{h \times w \times s\times s \times c \times c^{'}}$ and $W_{s} \in \mathbb{R}^{h \times w \times s\times s \times c \times c^{'}}$ is the shared weights. $\hat{\otimes}$ is the position-dependent convolution operation. The weights of our convolution are varied with the position of sliding window which is the main difference compared with traditional convolution.
	\subsubsection{Dynamic Weights Generation}
	We transform the generation of dynamic weights to the learning of coefficient vectors for several static kernels. We denote traditional convolution kernels as static kernels that share the weights among all inputs for all position to distinguish them from our dynamic kernels. In the coefficient generator, We use two traditional convolutional layers to predict the coefficients for different positions on the input. We set the kernel size to 3$\times$3 for the first layer and 1$\times$1 for the second layer. To reduce the computation, the channels of feature maps are squeezed to a quarter of original ones in the first layer. 
	We use $M$ static kernels denoted as $V=\{v_{i} | v_{i} \in \mathbb{R}^{s \times s \times c \times c^{'}}, i=1,..,M\}$. These kernels are defined as model parameters \cite{jia2016dynamic} as these in traditional convolutional layers which can only be changed after the back-propagation.
	Based on these static kernels, the coefficient generator $G_{dc}$ is used to regress the coefficients of different static kernels for different inputs and positions. We denote the output of $G_{dc}$ as $C_{d} \in \mathbb{R}^{h \times w \times M}$, which is both input-dependent and position-dependent. Then, the dynamic weights of filters at the position $[i,j]$ of the feature maps can be expressed as:
	\begin{align}
	\begin{split}
	W_{d}[i,j] = C_{d}[i,j,:] \times [v_{1},...,v_{M}]
	\end{split}
	\end{align}%
	
	Dynamic kernel can capture the variance on the different position of the feature maps. But it may suffer from insufficient training. Because  we split the gradients into several static kernels according to the coefficient vector, some kernels may be unable to receive enough updates considering the sparsity of the coefficient vector. Therefore, we introduce another static kernel whose coefficient is always 1 for all the positions on the feature map. To distinguish it from other static kernels, we denote it as shared weights. It is used to capture the common patterns among all positions of the input feature. They are defined as model parameters and updated by the back propagation only. Our combination of shared and dynamic weights decouples the perception of varied patterns. Although dynamic parts are theoretically capable of depicting the feature space, it will ease the task with both shared and dynamic parts considered. Taking the detection of cars from point cloud as an example, point density of different samples varies with the condition of the sample. However, they share some common features more or less because they are from the same kind of object. Therefore, the deployment of both position-independent shared filters and position-dependent dynamic filters is potential to achieve robust perception of changeable objects.  
	
	The total number of parameters of generated filters in our DD-Conv is $s \times s \times c \times c^{'} \times (M+1) + h \times w \times M$ where dynamic part contains $s \times s \times c \times c^{'} \times M + h \times w \times M$ parameters and static part contains $s \times s \times c \times c^{'}$ ones. Compared with DFN \cite{jia2016dynamic}, our DD-Conv greatly reduces the memory usage especially for feature maps with large size. The ratio of memory usage between our DD-Conv and DFN is:
	\begin{align}
	\begin{split}
	\frac{Mem_{our}}{Mem_{dfn}} & = \frac{s \times s \times c \times c^{'} \times (M+1) + h \times w \times M}{s \times s \times c \times c^{'} \times h \times w} \\
	& = \frac{M+1}{h \times w} + \frac{M}{s \times s \times c \times c^{'}}
	\end{split}
	\end{align}%
	As an example, with $s=3, c=128, c^{'}=128, h=248, w=216, M=3$, The memory consumed by the filters of DFN is 10,524 times than us.
	
	\subsection{Region Proposal Network}
	Our RPN adopts a dual-path design to further strength the representation ability and the adaptability to the variance on feature maps. In dynamic convolutional blocks (Dynamic ConvBlock), the last layer is replaced with our DD-Conv and other layers in the block adopt traditional convolution. The
	stride, number of convolutional layers and number of output channels of block 1 and block 2 are $[2; 6; 128]$, $[2; 6; 256]$. The layouts of Dynamic ConvBlock1 and ConvBlock2 are shown in Fig. \ref{ddblock}. Inspired by recent works \cite{lin2017feature,long2015fully,ronneberger2015u,hariharan2015hypercolumns} that combine multi-scale features to improve the detection performance, we use deconvolution to up-sample the features in shallow layers. All the features are up-sampled to the same size that is half of the size of inputs and their channels are set to 128. Features with different scales are concatenated to make the final detection. We use one traditional convolution layer as the detection head to make the final detection. On each position of the feature maps, we predict the results for two anchors whose angles are set to 0 and 90 degrees respectively.
	
	\subsection{Loss}
	We adopt similar loss design as \cite{lang2019pointpillars} which comprises three parts. The ground truth boxes and anchors are defined as $[x^{gt},y^{gt},z^{gt},w^{gt},l^{gt},h^{gt},\theta^{gt}]$ and $[x^{a},y^{a},z^{a},w^{a},l^{a},h^{a},\theta^{a}]$ respectively. The residual targets between ground truth and anchors are defined by:
	\begin{align}
	\begin{split}
	& \Delta x = \frac{x^{gt}-x^{a}}{d^{a}}, \Delta y = \frac{y^{gt}-y^{a}}{d^{a}}, \Delta z = \frac{z^{gt}-z^{a}}{h^{a}},\\ & \Delta w = log\frac{w^{gt}}{w^{a}}, \Delta l = log\frac{l^{gt}}{l^{a}}, \Delta h = log\frac{h^{gt}}{h^{a}}, \\ & \Delta \theta = sin(\theta^{gt}-\theta^{a}),
	\end{split}
	\end{align}%
	where $d^{a}=\sqrt{(w^{a})^{2}+(l^{a})^{2}}$. The localization loss is:
	\begin{align}
	\begin{split}
	\mathcal{L}_{loc} = \sum_{b \in (x,y,z,w,l,h,\theta)}SmoothL1(\Delta b).
	\end{split}
	\end{align}%
	Classification loss is defined as focal loss \cite{lin2017focal} as follows:
	\begin{align}
	\begin{split}
	\mathcal{L}_{cls} = -\alpha(1-p^{a})^{\gamma}logp^{a},
	\end{split}
	\end{align}%
	where $\alpha=0.25$, $\gamma=2$, $p^{a}$ is the classification probability. To distinguish the flipped boxes, a softmax loss $\mathcal{L}_{dir}$ is added to learn the direction. Therefore, the total loss for detection task is:
	\begin{align}
	\begin{split}
	\mathcal{L} = \frac{1}{N_{pos}}(\beta_{loc}\mathcal{L}_{loc}+\beta_{cls}\mathcal{L}_{cls}+\beta_{dir}\mathcal{L}_{dir}),
	\end{split}
	\end{align}%
	where $N_{pos}$ is the number of positive anchors and $\beta_{loc}=2.0$, $\beta_{cls}=1.0$ and $\beta_{dir}=0.2$.
	
	% To diverge the static kernels of our DD-Conv and make them orthogonal with each other as much as possible, we add a similarity loss using normalized dot products between different tensors as follows. When we use DD-Conv in our network, the similarity loss is added to the detection loss with a coefficient of 0.1.
	% \begin{align}
	% \begin{split}
	% \mathcal{L}_{sim} = \frac{1}{M}\sum_{i,j \in [1,...,M];i<j}\frac{\left|v_{i}\cdot v_{j}\right|}{\left\|v_{i}\right\|_2\cdot\left\|v_{j}\right\|_2}.
	% \end{split}
	% \end{align}%
	\begin{table*}
		\small 
		\begin{center}
			\caption{Performance of 3D and BEV detection results on the \emph{CAR} class of the KITTI validation set at the IoU of 0.7. ``R+L": RGB and LiDAR data. ``L": LiDAR data.}
			\label{tab:val}
			\begin{tabular}{c|c|c|ccc|ccc}
				\hline
				\multirow{2}{*}{Method} & \multirow{2}{*}{Modality} &
				\multirow{2}{*}{Stage} &
				\multicolumn{3}{c|}{3D AP} & \multicolumn{3}{c}{BEV AP}\\
				&&&\underline{Moderate} & Easy  & Hard 
				& \underline{Moderate} & Easy  & Hard \\
				
				\hline
				MV3D \cite{chen2017multi}  & R+L & Two & 62.68  & 71.29 & 56.56 & 78.10 & 86.55 & 76.67\\
				ContFuse \cite{liang2018deep}  & R+L & Two & 73.25 & 86.32  & 67.81 & 87.34 & {95.44}  & 82.43 \\
				AVOD-FPN \cite{ku2018joint}  & R+L & Two & 74.44 & 84.41  & 68.65 & -&-&-\\			
				F-PointNet \cite{qi2018frustum}  & R+L & Two & 70.92 & 83.76  & 63.65 & 84.02 & 88.16 & 76.44 \\	
				MMF \cite{liang2019multi}  & R+L &Two & 77.86 & 87.90 & 75.57 & 88.25 & \bf{96.66} & 79.60\\	
				PointRCNN  & L & Two  & 78.63 & 88.88  & 77.38 & -&-&-\\
				Part A$^{2}$ \cite{shi2019part} & L & Two & 79.47 & 89.47 & 78.54 & \bf{88.61} &90.42 & 87.31\\
				STD \cite{yang2019std} & L & Two& \bf{79.80} & \bf{89.70} & \bf{79.30} & 88.50 & 90.50 & \bf{88.10} \\
				\hline 	
				VoxelNet \cite{zhou2018voxelnet} & L & One& 65.46& 81.98  & 62.85 & 84.81 & 89.60  & 78.57\\
				SECOND \cite{yan2018second} & L & One & 76.48 & 87.43  & 69.10 & 87.07 & 89.96 & 79.66 \\
				Voxel-FPN \cite{wang2019voxel} & L &One &77.86 & 88.27 &75.84 &87.92 & 90.20 & 86.27 \\
				
				Ours & L & One  & \bf{78.25} &\bf{88.44} & \bf{76.03} & \bf{88.02} &  \bf{90.41}  & \bf{86.30}\\
				\hline
			\end{tabular}
		\end{center}
		\vspace{-0.2cm}
	\end{table*}
	\begin{table*}
		\small 
		\begin{center}
			\caption{Performance evaluation on KITTI 3D and BEV detection test set for \emph{CAR}. ``R+L": RGB and LiDAR data. ``L": LiDAR data.
			}
			\label{tab:testcar}
			\scalebox{0.9}{
				\begin{tabular}{c|c|c|c|ccc|ccc}
					\hline
					\multirow{2}{*}{Method} & \multirow{2}{*}{Modality} &
					\multirow{2}{*}{Stage} &\multirow{2}{*}{FPS}  & \multicolumn{3}{c|}{Car (3D)} & \multicolumn{3}{c}{Car (BEV)}\\
					&&&& \underline{Moderate} & Easy & Hard & \underline{Moderate} & Easy & Hard  \\
					\hline
					MV3D \cite{chen2017multi} & R+L & Two &2.8 & 63.63 & 74.97 & 54.00 & 78.93 & 86.62 & 69.80 \\
					AVOD-FPN \cite{ku2018joint} & R+L & Two &10& 71.76 & 83.07 & 65.73 & 84.82& 90.99 & 79.62 \\			
					F-PointNet \cite{qi2018frustum} & R+L &Two &5.9& 69.79 & 82.19  & 60.59 & 84.67 & 91.17  & 74.77  \\
					MMF \cite{liang2019multi} &R+L &Two &12.5 & 77.43 &88.40 &70.22 & 88.21 & 93.67 &81.99 \\
					PointRCNN \cite{shi2019pointrcnn} & L & Two &10 & 75.64 & 86.96 &70.70 &87.39 & 92.13 &82.72\\
					Part A$^2$ \cite{shi2019part} & L & Two & 12.5 & 78.49 & 87.81 & 73.51 &87.79 &91.70 &84.61\\
					STD \cite{yang2019std} & L & Two & 12.5 & \bf{79.71} & \bf{87.95} & \bf{75.09} & \bf{89.19} & \bf{94.74} & \bf{86.42} \\
					\hline 	
					ContFuse \cite{liang2018deep} & R+L &One &16.7& 68.78 & 83.68 & 61.67 & 85.35 & {\bf{94.07}} & 75.88 \\
					%					MMF \cite{liang2019multi} & RGB + LiDAR & 77.43 / 76.75 & 88.40 / 86.81 & 70.22 / 68.41 & 88.21 / 87.47 & 93.67 / 89.49 & 81.99 / 79.10\\	
					
					VoxelNet \cite{zhou2018voxelnet}& L &One & 4.4 &  65.11 & 77.47 &57.73 & 79.26 & 89.35 & 77.39  \\
					SECOND \cite{yan2018second} & L &One & 20 & 72.55& 83.34 & 65.82 & 83.77 & 89.39 & 78.59 \\
					PointPillars \cite{lang2019pointpillars} & L &One & 62 &74.31  & 82.58 & 68.99 & 86.56 & 90.07 & 82.81  \\
					Voxel-FPN \cite{wang2019voxel} & L & One & 50 & \bf{76.70} & \bf{85.64} & 69.44 & 87.21 & 92.75 & 79.82\\
					%			\hline 
					% 	PointRCNN \cite{shi2019pointrcnn} & LiDAR & 75.64 / 75.76 & {\bf86.96} / 85.94 & {\bf70.70} / 68.32  & {\bf87.39} / 85.68 & {\bf92.13} / 89.47 & 82.72 / 79.10  \\
					Ours & L &One &30& 76.24 & 85.30 & \bf{70.45} & {\bf87.25} & 90.87  & {\bf83.38} \\
					\hline
				\end{tabular}
			}	
		\end{center}
		\vspace{-0.3cm}
	\end{table*}
	\begin{table*}
		\small 
		\begin{center}
			\caption{Performance evaluation on KITTI 3D and BEV detection test set for \emph{CYCLIST}. ``R+L": RGB and LiDAR data. ``L": LiDAR data.
			}
			\label{tab:testcyc}
			\scalebox{0.9}{
				\begin{tabular}{c|c|c|ccc|ccc}
					\hline
					\multirow{2}{*}{Method} & \multirow{2}{*}{Modality}  &
					\multirow{2}{*}{Stage}  &\multicolumn{3}{c}{Cyclist (3D)} & \multicolumn{3}{c}{Cyclist (BEV)}\\
					&&&\underline{Moderate} & Easy & Hard & \underline{Moderate} & Easy & Hard  \\
					\hline
					AVOD-FPN \cite{ku2018joint} & R+L & Two & 50.55 & 63.76  & 44.93 & 57.12 & 69.39  & 51.09 \\			
					F-PointNet \cite{qi2018frustum} & R+L &Two & 56.12 & 72.27  & 49.01  & 61.37 & 77.26  & 53.78 \\
					PointRCNN \cite{shi2019pointrcnn} & L & Two & 58.82 & 74.96 & 52.53 & \bf{67.24} & \bf{82.56} & \bf{60.28}\\
					STD \cite{yang2019std} & L & Two & \bf{61.59} & \bf{78.69} & \bf{55.30} & 67.23 & 81.36 & 59.35\\
					\hline 		
					VoxelNet \cite{zhou2018voxelnet}& L &One & 48.36 &  61.22 &44.37 &54.76 &66.70 & 50.55 \\
					SECOND \cite{yan2018second} & L &One & 52.08 & 71.33  & 45.83  & 56.05& 76.50 & 49.45 \\
					PointPillars \cite{lang2019pointpillars} & L & One & 58.65 & {\bf77.10} & 51.92 & 62.73 & {\bf{79.90}} & 55.58  \\
					%			\hline 
					% 	PointRCNN \cite{shi2019pointrcnn} & LiDAR & 58.82 / 59.60 & 74.96 / 73.93  & 52.53 / 53.59 & {\bf67.24} / 66.77  & {\bf82.56} / 81.52 & {\bf60.28} / 60.78 \\
					Ours & L&One & {\bf59.54} & 75.43  & {\bf 53.37} & {\bf65.12} & 79.51  & {\bf{58.25}}  \\
					\hline
				\end{tabular}
			}	
		\end{center}
		\vspace{-0.3cm}
	\end{table*}
	\section{Experiments}
	We conduct our experiments on KITTI dataset \cite{geiger2012we}. Details of our network and experiments are demonstrated in Sec.\;\ref{5.1}. The comparison with other methods are shown in Sec.\;\ref{5.2}. In Sec.\;\ref{5.3}, ablation studies are conducted to analyze our proposed method.
	
	\subsection{Implementation Details}
	\label{5.1}
	During the voxelization stage, we first crop the whole point cloud within the range [[-39.68, 39.68], [0, 69.12], [-1, 3]]. The size of cell is set as [0.16, 0.16, 4].
	For the decomposable dynamic convolution, we use three static kernels ($M=3$) in all experiments unless explicitly stated in the ablation study.
	%We treat the representative tensors and shared weights as model parameters and enable network $G_{dc}$ for the generation of coefficients to reduce the computation and memory usage. 
	
	Following the split manner in \cite{chen2017multi}, the training set of KITTI dataset is divided into \emph{train} set (3712 images) and \emph{val} set (3769 images). We use an Adam optimizer to train our network. The learning rate is set as 0.0002 initially and decays by a factor of 0.8 for every 15 epochs. We train our network for 160 epochs with a batch size of 2 using a NVIDIA Titan V GPU card.
	
	\subsection{Results}
	\label{5.2}
	We evaluate our network on KITTI dataset for both validation set and the official test benchmark for \emph{car} and \emph{cyclist}. Table\;\ref{tab:val} shows our results on validation set for 3D and bird eye's view (BEV) detection. In Table\;\ref{tab:testcar} and Table\;\ref{tab:testcyc}, we show our results on the KITTI benchmark. Due to the evaluation policy change of KITTI benchmark on 08.10.2019, we show the results on KITTI benchmark of most methods with 40 recall positions except VoxelNet \cite{zhou2018voxelnet} because it only provides the data with 11 recall positions. However, most methods report their results with 11 recall positions on KITTI validation set. For a fair comparison with them, we also report the results with 11 recall positions on the validation set.
	
\begin{figure*}
		\begin{center}
			\includegraphics[scale=0.75]{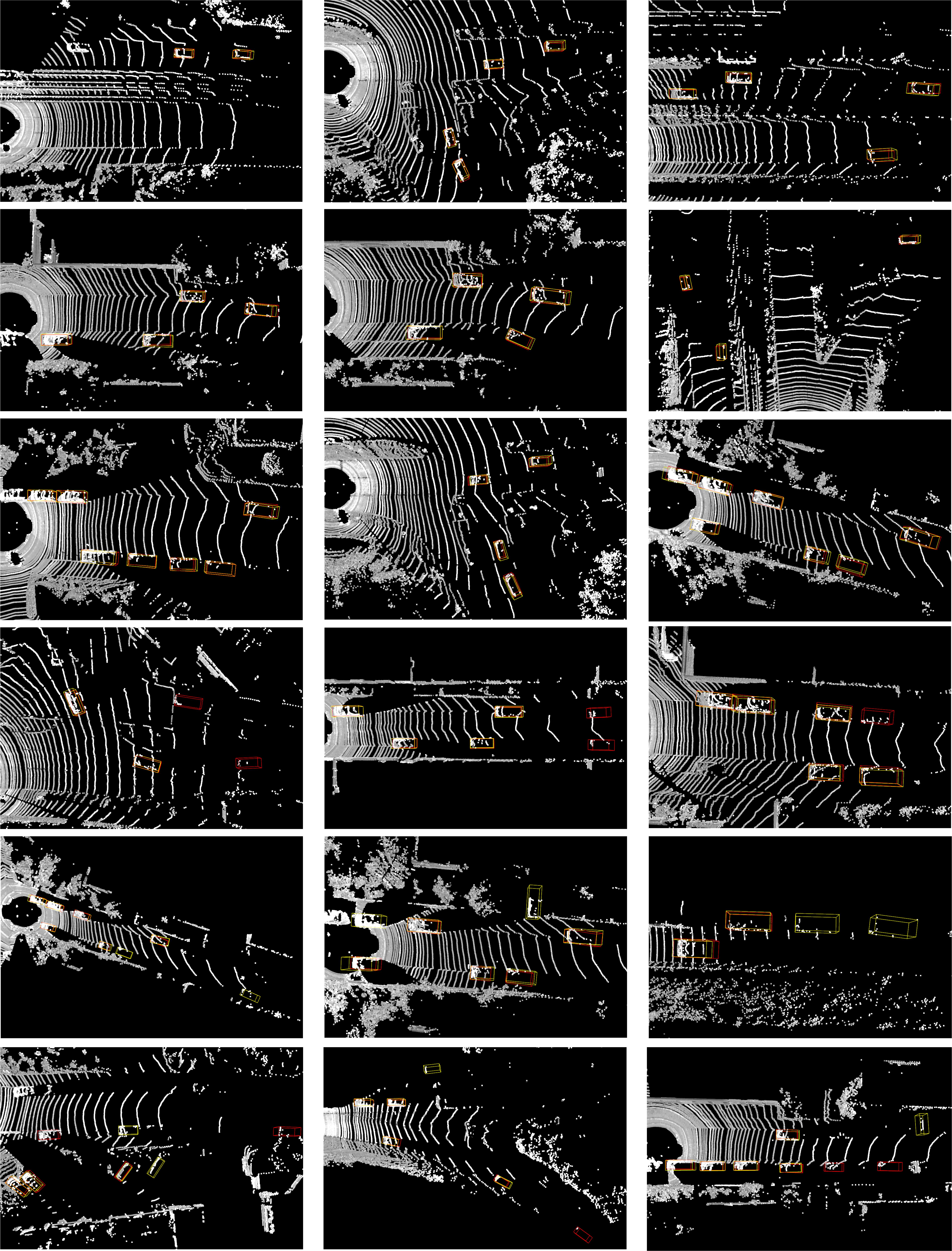}
		\end{center}
		\caption{The demos of 3D detection on KITTI validatioin set. }
		\label{fig:demo_3d}
	\end{figure*}

	On the validation set, we achieve dominant performance among all the one-stage voxelization-based methods on 3D detection and BEV detection tasks. For KITTI test benchmark, We get better results than all the single-scale one-stage methods by a clear margin. We outperform PointPillars \cite{lang2019pointpillars} for all the entries on the test set for \emph{car} class. We are 1.93, 2.72 and 1.46 points higher than PointPillars on the 3D detection task for \emph{car} under all the modes. Compared with VoxelFPN \cite{wang2019voxel} which use both multi-scale strategy and SSD head, our results are still competitive. We achieves better performance for all the entries on validation set and the detection for hard objects on test set. We visualize several demos of 3D detection in Fig. \ref{fig:demo_3d} on KITTI validation set. The first three rows show the objects that are correctly detected. The fourth row and sixth row give some missing cases in detection. They are usually caused by the far distance between the object and LiDAR sensor which makes the point cloud of the object very sparse. In the fifth row and sixth row, we gives some cases of false detection. They false detection is mainly caused by some confusing background such as wall and some cuboid objects. Besides, the missing labeling of KITTI dataset also leads to some false cases.
	\subsection{Ablation study}
	\label{5.3}
	% \begin{table}
	% 	\small 
	% 	\begin{center}
	% 		\scalebox{0.92}{
	% 			\begin{tabular}{c|ccc}
	% 				\hline
	% 				\multirow{2}{*}{Configure}  & \multicolumn{3}{c}{3D AP}\\
	% 				&\underline{Moderate} & Easy &Hard \\
	% 				\hline
	% 				DD-Conv (without shared part) & 76.84 & 86.86 & 70.24\\
	% 				DD-Conv (with shared part)  & \bf{77.29} & \bf{86.88} & \bf{72.88}  \\
	% 				\hline
	% 			\end{tabular}
	% 		}
	% 	\end{center}
	% 	\caption{Performance of DD-Conv with and without shared part on the \emph{car} class of the KITTI \emph{val} set.}
	% 	\label{tab:abl_share}
	% 	\vspace{-0.2cm}
	% \end{table}
	In this part, we analyze each component in our network. We make all the experiments on the \emph{car} class of  KITTI validation set. For our baseline network, we use the reproduced PointPillars \cite{lang2019pointpillars} which only considers pillar features and use traditional convolutional layers. Table\;\ref{tab:abl_cfg} shows the gains of performance by introducing the context features and our DD-Conv. To better evaluate our DD-Conv, we separately consider the influence of dynamic weights and shared weights. It's worth noticing that the experiments with only shared weights are equivalent to the baseline that uses traditional convolution. In our DD-Conv, the model adding shared part achieves better results than that with dynamic part only. The ablation study shows that our context features and DD-Conv can promote the performance of 3D detection whether being used individually or being combined together. With context features and DD-Conv, we are 2.39, 3.23 and 6.52 points higher than the baseline under \emph{moderate}, \emph{easy} and \emph{hard} mode for 3D detection. 
	\begin{figure}
		\begin{center}
			\includegraphics[scale=0.5]{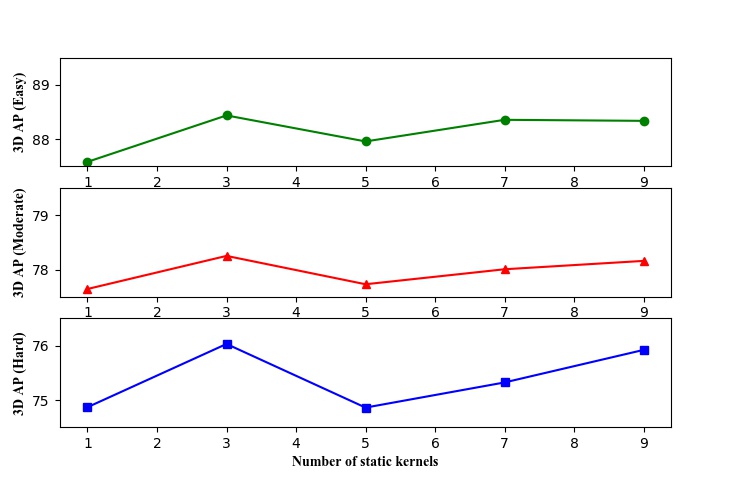}
		\end{center}
		\caption{The change of 3D AP over the number of representative tensors ($M$) on \emph{CAR} class of KITTI validation set. }
		\label{fig:AP_change}
	\end{figure}
	Fig. \ref{fig:AP_change} illustrates the variation of AP on 3D detection for \emph{car} class when we use different number of static kernels (denoted as $M$). When $M=3$, we achieve better results. The AP decreases more or less when we use more representative tensors in our dynamic convolution. This may be caused by our simple coefficient learning network $G_{dc}$. With only a bottleneck and an output layer, learning the relationship between too many tensors can be a hard task. Besides, with more static kernels, the coefficient tensor can be more sparse which is unfavourable to the updating of kernel weights. 
	
	To illustrate the effectiveness of our point-level context features and the dual-path RPN, we remove the DD-Conv layers in our network and compare our results with the multi-scale voxelization and FPN-based fusion method proposed in VoxelFPN \cite{wang2019voxel} which uses the same baseline (PointPillars \cite{lang2019pointpillars}) as us. To make a fair comparison, we change only one component in our CADNet once. The results are shown in Table\;\ref{tab:abl_ms}. To compare our point-level context features and the multi-scale features in VoxelFPN \cite{wang2019voxel}, we replace the point-level context features in our CADNet with multi-scale features. Other parts such as the pillar encoder and the dual-path RPN in the network are kept unchanged. We use two kinds of scales to extract multi-scale features and the larger scale is 3 times the scale of smaller one. To make the multi-scale features have same resolution as our point-level context, we up-sample the generated voxel features. The results of multi-scale features lags our point-level context by 0.55 points and 1.48 points for \emph{moderate} and \emph{easy} mode respectively, which shows the advantages of our voxelization method to generate the context for each pillar. Then, we compare the FPN-based fusion with our dual-path RPN. In this experiment, we use the point-level context features and change our dual-path RPN to FPN-based fusion network in VoxelFPN \cite{wang2019voxel}. From Table\;\ref{tab:abl_ms}, we can see that the performance drops by 0.83 points and 4.54 points respectively for \emph{moderate} and \emph{hard} mode. 
	\begin{table}
		\small 
		\begin{center}
			\caption{Performance of 3D object detectors with different configurations on the \emph{car} class of the KITTI \emph{val} set.}
			\label{tab:abl_cfg}
			\scalebox{0.92}{
				\begin{tabular}{ccc|ccc}
					\hline
					\multicolumn{3}{c|}{Configuration} & \multicolumn{3}{c}{3D AP (IoU=0.7)}\\
					context feature & dynamic & shared part &\underline{Moderate} & Easy & Hard\\
					\hline 
					& & \cmark & 75.86 & 85.21 & 69.51\\
					& \cmark & & 77.14 & 86.53 & 74.21\\
					\cmark &  & & 77.74 & 88.36 & 75.02\\
					& \cmark &\cmark & 77.57 & 86.88 & 75.59\\
					\cmark & \cmark &\cmark & \bf{78.25} & \bf{88.44} & \bf{76.03}\\
					\hline
				\end{tabular}
			}
		\end{center}
		\vspace{-0.2cm}
	\end{table}
	
	Table\;\ref{tab:abl_dfn} shows the comparison of different dynamic filters. Due to the huge memory usage of DFN \cite{jia2016dynamic}, it can only be used in the detection head of our network. To make a fair comparison, we only replace the layer in detection head with different dynamic filters and keep other layers unchanged. In our detection head, we take the input with size $[248,216,386]$ and output the detection results with shape $[248,216,20]$ (each position has 2 anchors and each anchor needs 7 box targets, 1 classification target and 2 direction targets). For Depth-aware convolution \cite{brazil2019m3d}, we set the number of bins to 27 like \cite{brazil2019m3d} which is claimed to have the best performance compared with other choices. Compared with DFN \cite{jia2016dynamic}, we reduce the number of parameters generated by the dynamic convolution layer from 413,544,960 to 23,160 while achieve comparable results. This greatly reduce the cost of integrating dynamic layers into existing models. Different from the predefined rows in depth-aware convolution layer \cite{brazil2019m3d}, our filters are much more flexible which can generate position-specific filters varying with input features at different positions. 
	
\begin{figure}
	\begin{center}
		\includegraphics[scale=0.38]{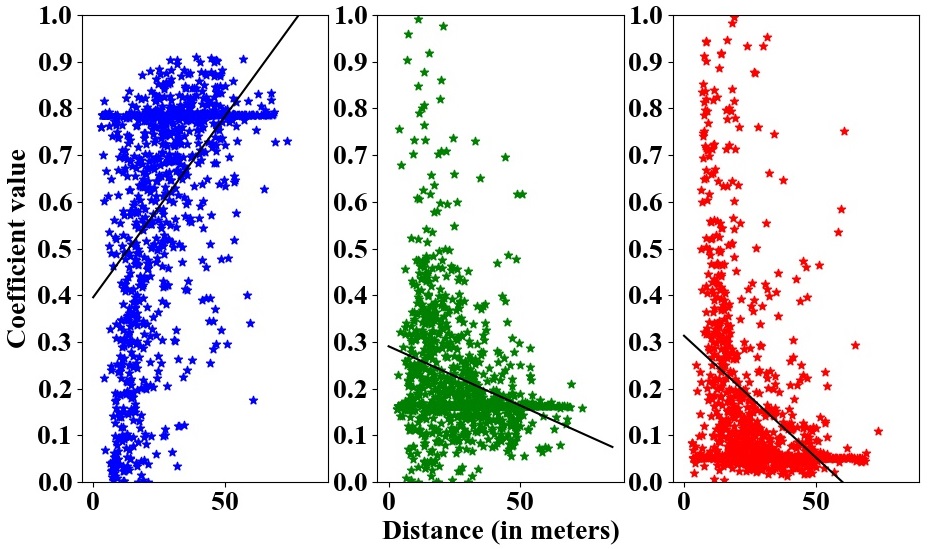}
	\end{center}
	\caption{Visualization of the relationship between coefficient vector of our DD-Conv in Dynamic ConvBlock0 and object's distance.}
	\label{fig:dis_coef0}
\end{figure}

\begin{figure}
	\begin{center}
		\includegraphics[scale=0.38]{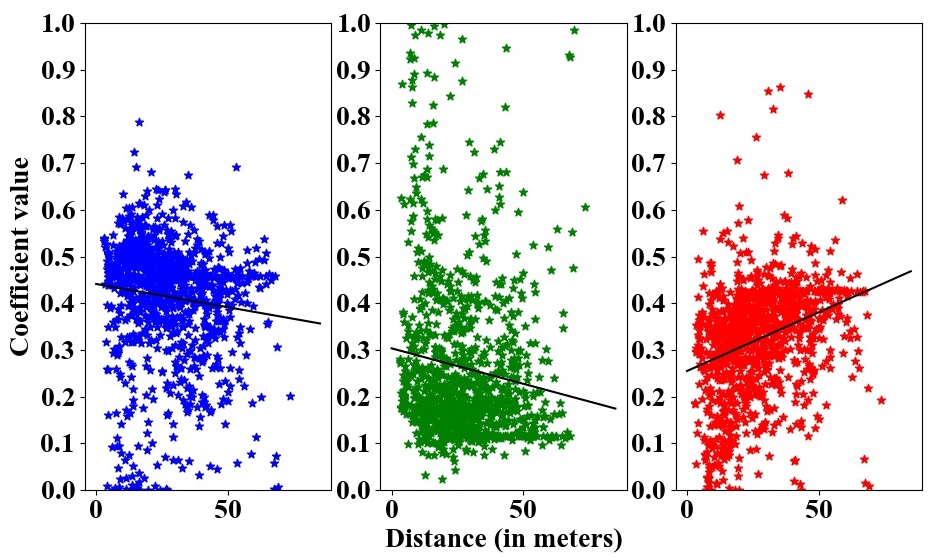}
	\end{center}
	\caption{Visualization of the relationship between coefficient vector of our DD-Conv in Dynamic ConvBlock1 and object's distance.}
	\label{fig:dis_coef1}
\end{figure}

\begin{figure}
	\begin{center}
		\includegraphics[scale=0.38]{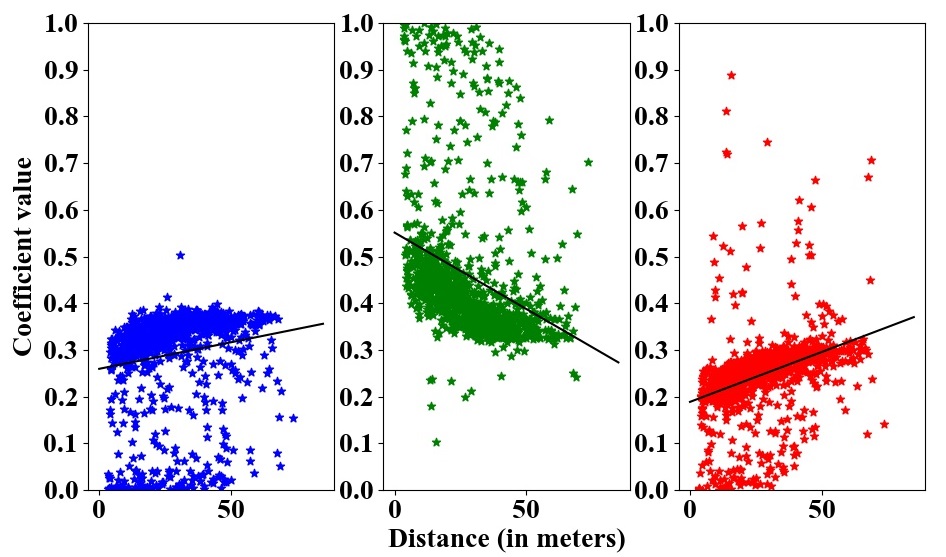}
	\end{center}
	\caption{Visualization of the relationship between coefficient vector of our DD-Conv in Dynamic ConvBlock2 and object's distance.}
	\label{fig:dis_coef2}
\end{figure}

\begin{figure*}
		\begin{center}
			\includegraphics[scale=0.6]{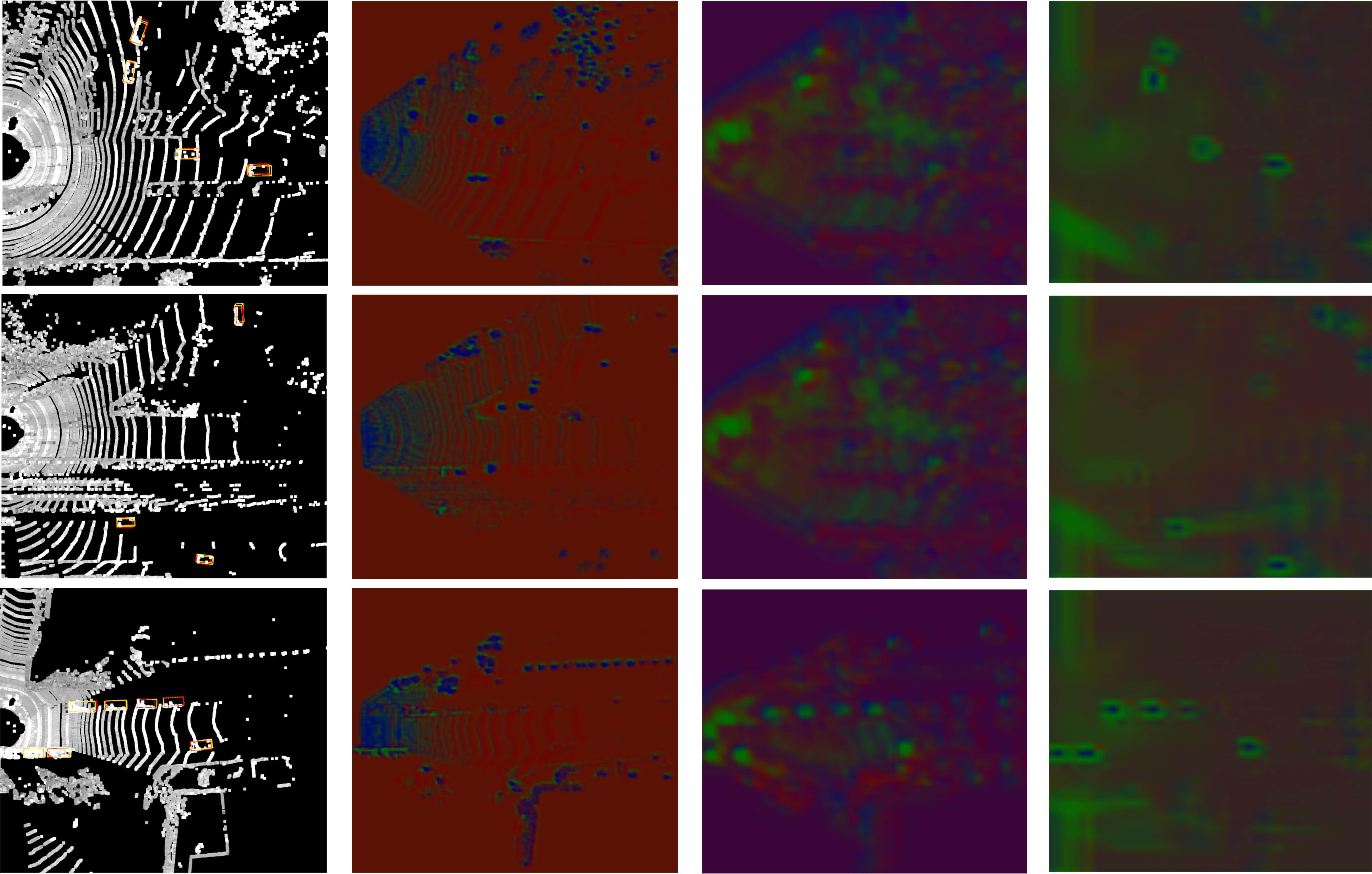}
		\end{center}
		\caption{Visualization of the coefficient vectors of our DD-Conv in Dynamic ConvBlocks. }
		\label{fig:coef_vis}
	\end{figure*}

	% Our decomposable dynamic convolution can be conveniently plugged into existing convolutional neural models without any extra change. We conduct experiments on two of the typical voxelzation-based models, PointPillars \cite{lang2019pointpillars} and SECOND \cite{yan2018second}. In our implementation, we reproduce these two network and replace traditional convolutional layers with our DD-Conv. The results ares shown in Table\;\ref{tab:abl_pp_second}. We witness a much higher promotion in PointPillars than SECOND. It is probably related to the usage of sparse convolution \cite{graham20183d} in SECOND which claims to better cope with sparse point cloud. 
	To better understand how these coefficients change for foreground objects, we analyze the variation of $M$ coefficients with the distance of the object. We take distance as a key property of object because it is highly related to the density and quality of points. We sample 300 LiDAR data from our validation set. The distance of every \emph{car} and  the corresponding coefficient vector are shown in Fig. \ref{fig:dis_coef0}, Fig. \ref{fig:dis_coef1} and Fig. \ref{fig:dis_coef2}. We analyze the DD-Conv layers in different Dynamic ConvBlocks. Fig. \ref{fig:dis_coef0}, Fig. \ref{fig:dis_coef1} and Fig. \ref{fig:dis_coef2} show the statistics for the DD-Conv in block 0, block 1 and block 2 respectively. We can see that coefficient vector varies with the distance of the object, which means our network learns to generate different convolutional kernels for the objects with different distances. It verifies our intuition to handle highly changeable objects with different filter in a dynamic manner. Besides, the changing trend of coefficients over distance in multiple blocks is distinguishable which indicates that the variance of input features are different in different layers. Shallow layers usually can reflect the variance of texture cues and deep layers can show the change of semantic cues. 
	In addition to the above analysis of our DD-Conv on the foreground objects, we also visualize the coefficient vector of our DD-Conv on the whole scene as shown in Fig. \ref{fig:coef_vis}. The visualization results of the coefficients for our DD-Conv in block 0, block 1 and block 2 are shown in the second, third and fourth columns respectively. As we can see, in shallow layer (block 0), our DD-Conv can distinguish the difference between some easy patterns such as empty regions, points in line and points in cluster. In deep layers, our DD-Conv can gradually respond differently to positive and negative samples. This kind of difference may help the network to decompose a task into smaller sub-tasks and solve them with different filters.
	\begin{table}
		\small 
		\begin{center}
			\caption{Comparison with detectors using different voxelization strategy and fusion method on the \emph{car} class of the KITTI \emph{val} set.}
			\label{tab:abl_ms}
			\scalebox{0.92}{
				\begin{tabular}{c|ccc}
					\hline
					\multirow{2}{*}{Method}  & \multicolumn{3}{c}{3D AP}\\
					&\underline{Moderate} & Easy &Hard \\
					\hline
					PointPillars & 75.86 & 85.21 & 69.21\\
					Multi-scale features & 77.19 & 86.88 & 75.00 \\
					FPN-based fusion & 76.91 & 87.55 & 70.48 \\
					Ours & \bf{77.74} & \bf{88.36} & \bf{75.02}  \\
					\hline
				\end{tabular}
			}
		\end{center}
		\vspace{-0.2cm}
	\end{table}
	\begin{table}
		\small 
		\begin{center}
			\caption{Performance of 3D object detectors with different dynamic filters on the \emph{car} class of the KITTI \emph{val} set.}
			\label{tab:abl_dfn}
			\scalebox{0.92}{
				\begin{tabular}{c|ccc|c}
					\hline
					\multirow{2}{*}{Method}  & \multicolumn{3}{c|}{3D AP} & \multirow{2}{*}{Parameters}\\
					&\underline{Moderate} & Easy &Hard &\\
					\hline
					Baseline & 75.86 & 85.21 & 69.21& -\\
					DFN\cite{jia2016dynamic}  & 77.02 & {\bf87.23} & 70.10 & 413,544,960 \\
					Depth-aware \cite{brazil2019m3d} & 76.40 & 84.61 & 69.63 & 208,440\\
					Ours & \bf{77.06} & 86.40 & \bf{70.16} & \bf{23,160} \\
					\hline
				\end{tabular}
			}
		\end{center}
		\vspace{-0.2cm}
	\end{table}

	\section{Conclusion}
	In this paper, we propose a strong one-stage 3D detector to tackle the varied density of point cloud. We introduce context features for each voxelized pillars to capture the variance of density and design a decomposeable dynamic layer to adapt to the change of local features. The proposed context features have a larger receptive field which can help to better describe the variance of local features. They also provide more diverse features to the following RPN. 
	Our dynamic convolutional layers decompose convolution into shared and dynamic parts. We innovatively take dynamic filtering as the learning of representative tensors and the combination of them. This insight helps us greatly reduce the number of parameters and ease the difficulty to fit varying features in point cloud. We take our experiments on KITTI dataset and achieve competitive performance compared with other voxelization-based methods.
	
	{\small
		\bibliographystyle{ieee_fullname}
		\bibliography{egbib}
	}
	
\end{document}